\documentclass[lettersize,journal]{IEEEtran}

\usepackage{amsmath,amsfonts}
\usepackage{float}
\usepackage{algorithm}
\usepackage{algpseudocode}
\usepackage{array}
\usepackage{multirow}
\usepackage{booktabs}
\usepackage{arydshln}
\usepackage{colortbl}
\usepackage{tcolorbox}
\usepackage{enumitem}
\usepackage{bbm}
\usepackage{textcomp}
\usepackage{url}
\usepackage{graphicx}
\usepackage{cite}

\definecolor{drssbg}{RGB}{240,248,255}
\begin{document}

\title{IntPro: Context-Aware Intent Inference via Selective Retrieval from User History}

\author{Guanming Liu, Meng Wu, Peng Zhang, Yu Zhang, Yubo Shu, Xianliang Huang, Kainan Tu, Ning Gu, Liuxin Zhang, QianYing Wang, and Tun Lu%
\thanks{Guanming Liu, Peng Zhang, Yubo Shu, Xianliang Huang, Kainan Tu, Ning Gu, and Tun Lu are with Fudan University, Shanghai, China (e-mail: gmliu24@m.fudan.edu.cn; zhangpeng\_@fudan.edu.cn; 20110240032@fudan.edu.cn; huangxl21@m.fudan.edu.cn; kntu25@m.fudan.edu.cn; ninggu@fudan.edu.cn; lutun@fudan.edu.cn).}%
\thanks{Meng Wu, Yu Zhang, Liuxin Zhang, and QianYing Wang are with Lenovo Group Ltd, Beijing, China (e-mail: wumeng5@lenovo.com; zhangyu29@lenovo.com; zhanglx2@lenovo.com; wangqya@lenovo.com).}}

\maketitle

\begin{abstract}
Accurately understanding user intent in context is essential for large language models (LLMs) to provide effective assistance. However, observed behavior may support multiple plausible intents, making the current context alone insufficient for identifying the user's underlying goal. A user's interaction history may contain recurring context-to-intent patterns that help disambiguate the current intent. Leveraging such historical patterns requires deciding whether to invoke retrieval for the current context and, if so, forming queries that express plausible intents rather than relying on contextual similarity alone. We present IntPro, a proxy agent that learns to either infer intent directly from the current context or generate candidate intent explanations to retrieve intent-aligned records from the user's history. We design intent explanations as a shared intent-oriented natural-language form that expresses the observed action, its functional goal, and a cue-to-intent rationale. Label-conditioned explanations represent historical records, while candidate explanations generated from the current context serve as retrieval queries in the same form. IntPro first learns direct and retrieval-conditioned inference through Direct-and-Retrieval supervised fine-tuning (SFT). IntPro then uses multi-turn GRPO with a retrieval-decision reward to jointly optimize intent prediction and selective retrieval mechanism. Experiments across three scenarios and three model backbones demonstrate that IntPro consistently improves context-aware intent inference.
\end{abstract}


\section{Introduction}

Large language models have shown broad potential as assistants in scenarios including knowledge acquisition, communication, and online content engagement. Effective assistance in these settings depends on identifying the user's intended goal from the available interaction context. In practice, however, an observed action often does not state that intention explicitly. For example, a highlighted technical term may correspond to a request for its definition, a comparison with related terms, or an account of the concept's origin and evolution. An ambiguous utterance or a social-media post can likewise support several plausible intents \cite{tong2022context,prakash2024unified}. We refer to this mismatch as the \emph{Intent Articulation Gap}: the observed action and its context may support multiple plausible intents without identifying which goal the user currently pursues. The situational context provides immediate cues for intent inference, while the same user's history may reveal recurring relations between contextual cues and intended goals. The challenge is therefore to determine when the current context requires retrieving such historical patterns.

Existing research has largely studied context-aware intent inference and retrieval-based personalization as separate problems. Intent detection models commonly predict a label under a predefined ontology from observed behavior and context \cite{dubiel2024device,feng2025improving}. Such models generally use the observed behavior and context as evidence for the current prediction. Personalized memory and user-history retrieval methods instead store or retrieve user-associated records to support personalized outputs \cite{zhong2023memorybank,packer2023memgpt,chhikara2025mem0,salemi2023lamp,10.1145/3696410.3714850}. In intent inference, however, the current intent is latent when a retrieval query is formed, so similarity computed directly from observed actions or full contexts provides only indirect evidence that a historical record expresses the same intent. Combining intent inference with personalized retrieval therefore requires deciding, for each context, whether retrieval is needed and, when it is, constructing an intent-oriented query before the current intent has been inferred.

\begin{figure}[t]
    \centering
    \includegraphics[width=\columnwidth]{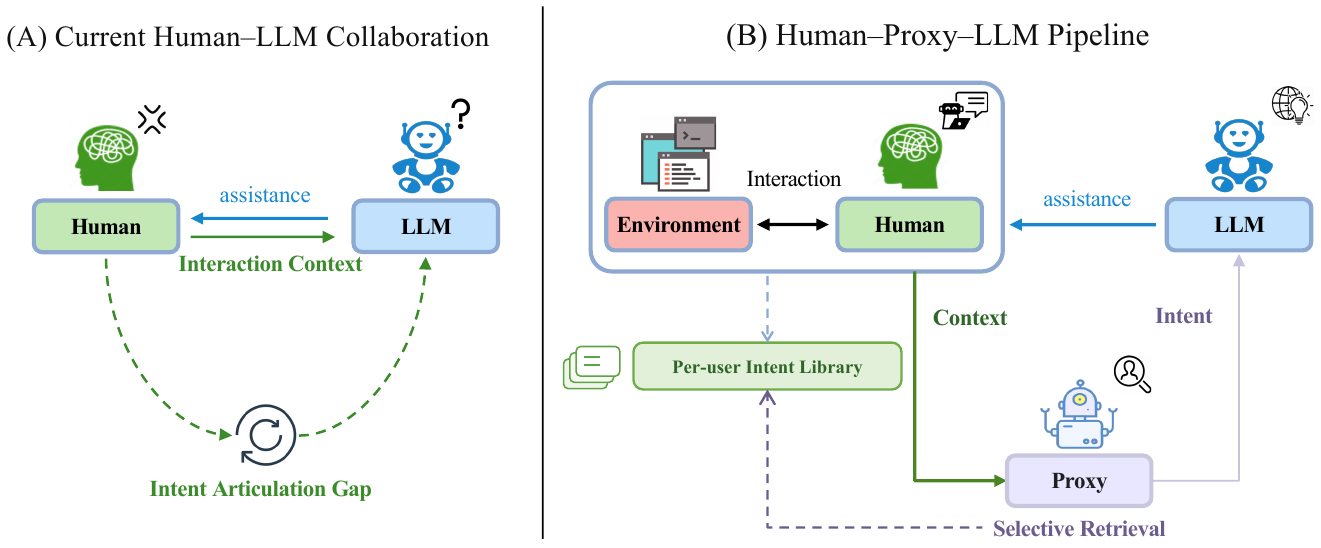}
    \caption{Conceptual comparison of current Human--LLM collaboration and the Human--Proxy--LLM pipeline. The proxy chooses between direct inference from the current context and retrieval-conditioned inference using the per-user intent library, then supplies the inferred structured intent to the downstream LLM.}
    \label{motivation}
    \end{figure}

We formulate this problem within the Human--Proxy--LLM pipeline illustrated in Figure~\ref{motivation} \cite{shu2024rah,10.1145/3696410.3714850,chen2025cpo}. Given the current context and access to a per-user intent library, the proxy either completes intent inference directly or invokes retrieval and uses the returned same-user records to complete the inference. It then outputs an intent label together with an intent explanation. The label identifies the predicted category, while the explanation describes how the observed behavior and context support that intent and can reflect patterns recorded in the same user's history when retrieval is used. The downstream LLM can then use the predicted intent and its explanation for task-specific assistance. This formulation assigns user-history retrieval and intent inference to the proxy while keeping downstream generation task-specific.

We propose IntPro, an intent-centric proxy agent that learns when and how to retrieve from a per-user intent library containing historical contexts, annotated intents, and natural-language intent explanations. Each historical record is represented for retrieval by a label-conditioned intent explanation that integrates the observed action, its functional goal, and a cue-to-intent rationale. Given a new context, IntPro either infers intent directly or generates candidate intent explanations from that context. These candidates share the same intent-oriented natural-language form as the historical explanations and serve as retrieval queries. IntPro uses the retrieved records together with the current context to infer the structured intent.

To realize this capability, we establish a training paradigm consisting of a supervised fine-tuning phase and a
reinforcement learning phase. In the supervised fine-tuning phase, we teach both inference paths from constructed multi-turn demonstrations. Direct trajectories produce the intent prediction from the current context, while retrieval-conditioned trajectories additionally generate candidate intent explanations, issue a retrieval call, and incorporate the returned records. We term this stage Direct-and-Retrieval SFT. In the reinforcement learning phase, multi-turn GRPO \cite{guo2025deepseek} lets the proxy sample its own trajectories following either path and optimizes intent inference together with the retrieval behavior, using a verifiable accuracy reward and a retrieval-decision reward. The latter adjusts the decision credit assigned to direct and retrieval-conditioned outcomes based on the group's prediction success rate, with both evaluated against final intent correctness.

We evaluate IntPro on Highlight-Intent, MIntRec2.0, and Weibo Post-Sync, spanning reading, dialogue, and social-media contexts. Across three model backbones, IntPro consistently improves context-aware intent inference over Naive GRPO. Further analyses show higher Accuracy and Macro-F1 under selective retrieval policy than under either fixed policy, while intent explanations support more intent-aligned retrieval than actions or full contexts in both historical-record representation and query formation.

The primary contributions are:
\begin{itemize}
\item We formulate context-aware intent inference with selective retrieval from user history and propose IntPro, a proxy agent that, for each context, chooses between direct inference and retrieval-conditioned inference over a per-user intent library.
\item We design intent explanations as a shared intent-oriented natural-language form in which label-conditioned explanations represent historical records and candidate explanations generated from the current context serve as retrieval queries. Direct-and-Retrieval SFT teaches IntPro both inference paths, while multi-turn GRPO with a retrieval-decision reward trains IntPro to decide whether to invoke retrieval.
\item We evaluate IntPro across three scenarios and three model backbones. Controlled comparisons show consistent intent-inference gains over Naive GRPO, and policy and retrieval analyses demonstrate the effectiveness of selective retrieval and intent-aligned retrieval.
\end{itemize}

\section{Related Work}
\label{sec:related}

\subsection{Context-Aware Intent Understanding}

Intent understanding aims to infer the latent goal behind observed user behavior. Context constrains this mapping because the same behavior can express different goals across situations, yet the current context can still support more than one plausible intent. Prior work studies how dialogue history and situational cues reduce ambiguity in context-dependent intent inference \cite{tong2022context,prakash2024unified}. Domain-adaptation methods transfer intent models across domains with limited labeled data \cite{atuhurra2024domain}, while multimodal and out-of-scope settings broaden the observations and intent spaces considered \cite{zhang2024mintrec}. Together, these approaches improve intent prediction from the available interaction evidence.

Recent work applies LLMs to intent detection through natural-language instructions and generative or zero-shot classification \cite{qin2025divide,arora2024intent}, and uses them to generate intent-aware information-seeking dialogues as additional training data \cite{askari2025solid}. Complementary studies examine LLM-based intent recognition and user satisfaction \cite{bodonhelyi2024user} and develop lightweight LLMs for on-device intent prediction \cite{dubiel2024device}. Intent inference nevertheless remains commonly formulated as predicting a label under a predefined ontology \cite{schick-schutze-2021-exploiting}. Open-intent discovery relaxes the closed-set assumption \cite{vedula2019openintent}, while rationalized prediction associates a decision with supporting evidence or an explanation \cite{deyoung-etal-2020-eraser}. Current intent-detection methods commonly incorporate interaction history directly into the prediction context without modeling retrieval from a per-user intent library as part of intent prediction.

\subsection{User Representations in Personalized LLMs}

Personalized LLMs incorporate user-specific information through model adaptation or external user representations. Population-to-individual tuning adapts pretrained language models for on-device user intent prediction \cite{gong2024population}. Other approaches compress interaction histories into user embeddings that can be injected into prompts or used to contextualize an LLM \cite{10.1145/3701716.3715463,doddapaneni2024user}. These methods learn user-conditioned predictions or compact user representations.

External memory and retrieval methods use user-associated records stored outside shared model parameters. Long-term memory systems store and retrieve information across conversations \cite{zhong2023memorybank,packer2023memgpt,chhikara2025mem0}, while recent work studies memory construction, retrieval, and reflective management for personalized dialogue agents \cite{pan2025secom,tan2025prospect}. Personalized language-model benchmarks such as LaMP retrieve user records to support user-specific outputs \cite{salemi2023lamp}, and retrieval systems incorporate author features and contrastive examples to represent user histories \cite{yazan2025improving}. These methods make user-specific information available to language models or maintain a user representation. For intent inference, the usefulness of user-history retrieval varies with the current context, motivating a per-context choice between direct and retrieval-conditioned inference.

\subsection{Training Retrieval-Augmented Agents}

Training an agent to retrieve requires representing the search action and its returned evidence within the model trajectory. ReAct represents reasoning and acting in a shared trajectory \cite{yao2023react}, while Toolformer learns where API calls are useful \cite{schick2023toolformer}. Subsequent work examines agent--computer interfaces, hybrid model architectures, and tool-use data collected across tasks \cite{yang2024sweagent,shao2025division,shi2025tool}. This broader tool-use literature establishes how retrieval calls and responses can enter an agent trajectory.

Search is one instance of model-initiated tool use. Open-RAG adaptively determines when external evidence is needed and improves how open-source language models reason over that evidence \cite{islam2024open}; Search-and-Refine interleaves search with explicit evidence refinement during reasoning \cite{shi2025search}; and Search-R1 uses reinforcement learning to train models to issue multiple search queries while reasoning \cite{jin2025search}. These methods use external evidence for knowledge-intensive question answering. Hypothetical Document Embeddings (HyDE) addresses query construction for zero-shot dense retrieval by generating a hypothetical document from the input query and using its embedding to retrieve similar corpus documents \cite{gao-etal-2023-precise}. IntPro uses a related generate-then-retrieve intuition for intent inference: candidate intent explanations generated from the current context serve as queries over label-conditioned intent explanations of historical records in a bounded, per-user intent library.

Learning when to call a tool introduces a credit-assignment problem between the action and the final task outcome. Supervised tool trajectories can establish call syntax and multi-turn behavior, while reinforcement learning can connect intermediate actions to task outcomes. StepTool and ToolRL attach learning signals to tool-use decisions \cite{yu2024steptool,qian2025toolrl}, and SWiRL optimizes multi-step tool trajectories \cite{goldie2025swirl}. Group-relative optimization offers a value-model-free recipe for reasoning post-training \cite{shao2024deepseekmath,guo2025deepseek}. A final intent reward alone leaves the retrieval decision underdetermined because it assigns the same credit to a correct label reached through either path and the same penalty to a failure on either path. Motivated by these works, IntPro therefore uses a retrieval-decision reward to learn when to invoke user-history retrieval during intent inference.

\section{Problem Formulation}
\label{sec:problem}
\subsection{Context-Aware Intent Inference}
\label{sec:context_intent}

Context is commonly understood as information that characterizes the circumstances surrounding an entity or action \cite{dey2001understanding}. In this work, the current context for intent inference combines the user's observable behavior with the interaction environment in which it occurs. We denote the interaction environment as $E$, the user's observable action or utterance as $A$, and the current context available for intent inference as $C=(E,A)$. This context constrains the interpretation of the action but may not uniquely determine its task-level intent.

Intent represents the underlying goal or purpose that motivates a user's action or query within a specific context. Given $C$, the proxy predicts a structured intent
\begin{equation}
    \hat{\mathcal{I}}=(\hat{\ell},\hat{e}),
\end{equation}
where $\hat{\ell}$ is a label in the task ontology and $\hat{e}$ is the corresponding intent explanation. We define this explanation as a structured natural-language representation:
\begin{equation*}
\begin{aligned}
\hat{e}
&=\operatorname{Compose}
\left(\hat{a}^{\star},\hat{g},\hat{r}\right),\\
\hat{a}^{\star}
&=\operatorname{Quote}(A),\qquad
\hat{g}=\operatorname{Goal}(C,\hat{\ell}),\\
\hat{r}
&:s(C,A)\Rightarrow\hat{\ell},
\end{aligned}
\end{equation*}
where $\hat{a}^{\star}$ retains the critical observable action, $\hat{g}$ states the functional goal that this action expresses in the current context, and $\hat{r}$ explains why a salient source-level cue $s(C,A)$ supports the predicted intent. $\operatorname{Compose}$ generates one natural-language explanation that integrates these three components.

This design combines source grounding with intent-oriented abstraction. Retaining the original action makes each explanation traceable to the utterance, post, or highlighted span from which it was inferred. The functional goal abstracts that concrete behavior into task-level intent semantics, while the cue-to-intent rationale explains how a salient source-level cue supports the inferred intent. The resulting intent explanation preserves its source evidence while providing a reusable representation for retrieval and the structured intent output.

\subsection{Per-User Intent Library}
\label{sec:history_library}

When the current context underdetermines intent, IntPro can retrieve from a per-user intent library that organizes the user's prior interaction records. For user $u$, the intent library $\mathcal{DB}_u^{\mathrm{lib}}$ is constructed from the interaction records $H_u^{\mathrm{lib}}$ assigned to the library split. Each library is confined to its own user's records, so inferring user $u$'s intent never accesses other users' interaction histories.

Each intent library entry
\begin{equation}
    d=(u,C,\ell,e)
\end{equation}
represents a historical interaction, pairing the context $C$ with an annotated intent label $\ell$ and a natural-language intent explanation $e$ describing how the context expresses that intent. The intent explanation $e$ is constructed from the historical context and its annotated intent and serves as the entry's label-conditioned retrieval representation. At inference time, IntPro generates candidate intent explanations in the same form from the current context $C$ and uses them as queries over the label-conditioned intent explanations of historical entries. Section~\ref{sec:data} details query generation and retrieval.

\subsection{Direct and Retrieval-Conditioned Inference}
\label{sec:inference_paths}

Because user history has context-dependent utility, the proxy can infer the structured intent through either direct inference from the current context alone or retrieval-conditioned inference that additionally uses same-user records retrieved from the intent library. The two paths are defined as follows:
\begin{equation}
\hat{\mathcal{I}}=
\begin{cases}
\operatorname{IntPro}(C),
& \text{direct inference},\\
\operatorname{IntPro}(C,\mathcal{R}),
& \text{retrieval-conditioned inference}.
\end{cases}
\label{eq:inference_paths}
\end{equation}
Equation~\ref{eq:inference_paths} distinguishes the information available to the two inference paths. Direct inference uses only the current context $C$, whereas retrieval-conditioned inference generates queries from $C$, invokes retrieval over the per-user intent library, and uses the returned same-user records $\mathcal{R}$, whose intent explanations include cue-to-intent rationales grounded in the historical contexts, to produce the final structured intent. Section~\ref{sec:data} details this retrieval procedure. The retrieval decision selects between these paths, both of which output a structured intent that can be passed to a downstream LLM for task-specific generation.

\section{Data Construction}
\label{sec:data}

\begin{figure}[t]
\centering
\includegraphics[width=0.5\textwidth]{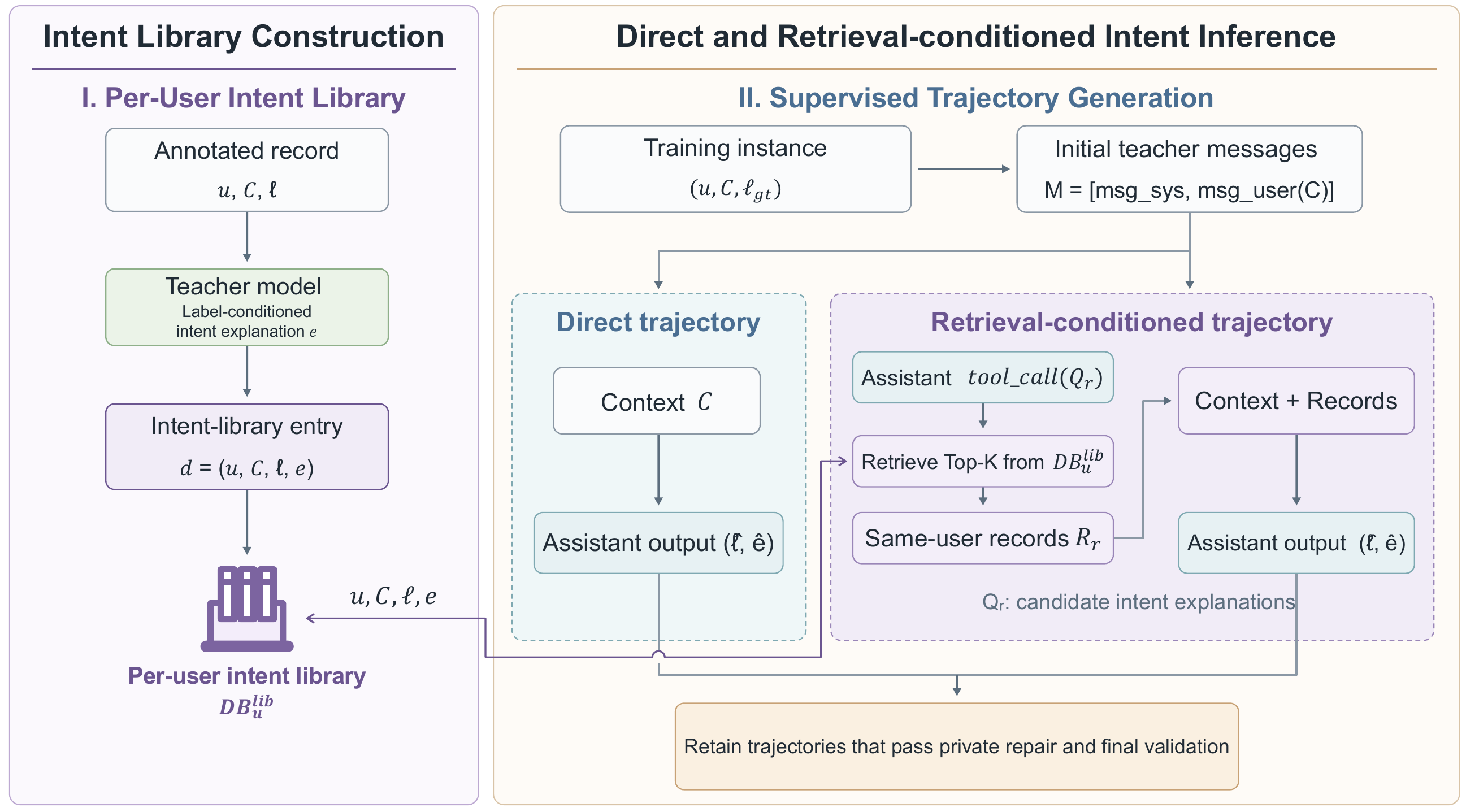}
\caption{Construction of the per-user intent library and supervised trajectories. Historical records are represented by label-conditioned intent explanations, while candidate intent explanations generated from $C$ serve as retrieval queries. Ground truth supports label validation and final teacher target construction.}
\label{data_rollout}
\end{figure}

Data construction defines the retrieval procedure and produces the two artifacts consumed by IntPro: a fixed per-user intent library and validated trajectories for direct and retrieval-conditioned inference. The library stores the historical records available to retrieval, while the trajectories demonstrate either a direct answer or a retrieval call followed by inference over the returned records. We use Qwen3-30B-A3B as the teacher model for both intent-library construction and supervised trajectory generation. Figure~\ref{data_rollout} summarizes how these artifacts are constructed.

\subsection{Intent Library Construction and Retrieval}

Section~\ref{sec:context_intent} defines an intent explanation that retains its source action, abstracts the action's functional goal, and includes a cue-to-intent rationale. During library construction, the teacher model transforms each annotated record $(u,C,\ell)$ into an entry $d=(u,C,\ell,e)$ by generating the corresponding label-conditioned, source-grounded intent explanation $e$. During retrieval, the proxy generates candidate intent explanations for a new context in the same form, enabling direct matching against the stored explanations. Multiple candidates allow the proxy to express alternative hypotheses about how the same context may indicate an intent. The returned records then condition the final prediction of $(\hat{\ell},\hat{e})$.

Let \(Q=\{q^{(j)}\}\) contain one or more candidate intent explanations used as retrieval queries, and let \(\phi(\cdot)\) denote BGE-M3 \cite{chen2024m3}. For each \(q^{(j)}\in Q\), let \(L_j\) be the Top-\(K\) records from the same-user library ranked by the cosine similarity between \(\phi(q^{(j)})\) and the stored representation \(\phi(e(d))\). For each record \(d\in\bigcup_j L_j\), we define its retrieval frequency and conditional mean similarity as
\begin{equation*}
f(d)=\sum_j\mathbbm{1}[d\in L_j],
\end{equation*}
\begin{equation*}
\bar{s}(d)
=
\frac{1}{f(d)}
\sum_{j:d\in L_j}
\cos\!\left(\phi(q^{(j)}),\phi(e(d))\right).
\end{equation*}
The mean similarity is computed only over the query-specific lists containing \(d\). We rank records in descending lexicographic order by \(\bigl(f(d),\bar{s}(d)\bigr)\): retrieval frequency counts how many candidate-specific Top-\(K\) lists contain a record, prioritizing records that recur across more candidates, and conditional mean similarity orders records with the same frequency. The final Top-\(K\) unique records form the retrieval result \(\mathcal{R}\).

\subsection{Training Trajectory Generation}

With the intent library fixed, the second construction stage generates the trajectories consumed by supervised fine-tuning.

\begin{figure*}[t]
\centering
\includegraphics[width=0.9\textwidth]{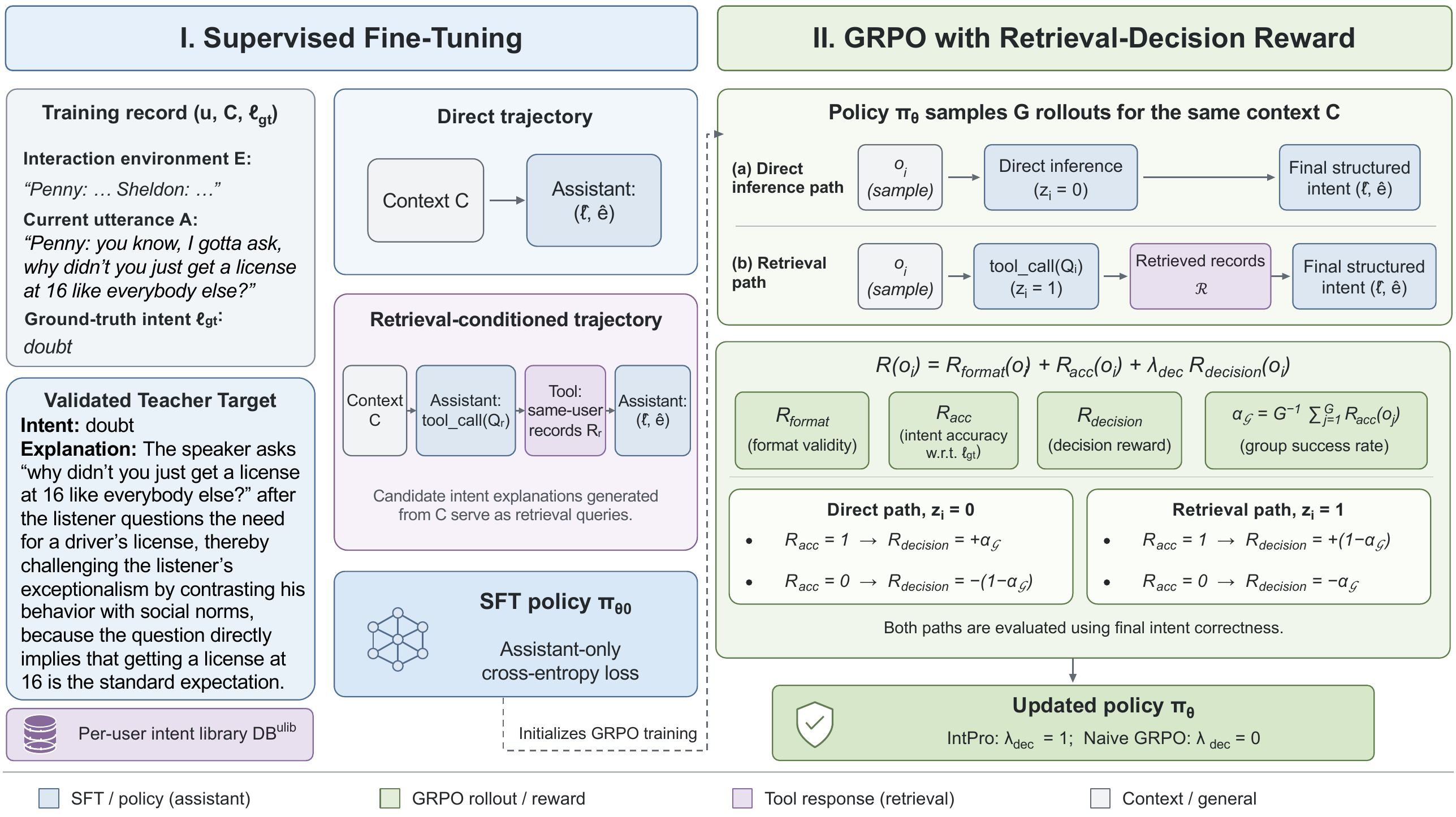}
\caption{Two-stage training of IntPro. SFT learns direct and retrieval-conditioned inference from validated teacher trajectories; GRPO then samples complete trajectories following either path and uses final intent correctness to optimize intent prediction and the retrieval decision.}
\label{training-pipeline}
\end{figure*}

\begin{algorithm}[H]
    \footnotesize
    \caption{Supervised Intent-Inference Trajectory Generation}
    \label{alg:contrastive_multiturn}
    \begin{algorithmic}[1]
    \Require
        Instance $(u,C,\ell_{gt})$, per-user intent library $\mathcal{DB}_u^{\mathrm{lib}}$, teacher LLM, trajectory-turn budget $I_{max}=3$, private repair budget $B=6$
    \Ensure
        Validated supervised trajectory or \textsc{Discard}
    
    \State $M \leftarrow [msg_{sys}, msg_{user}(C)]$
    \State $\mathcal{R} \leftarrow \emptyset$
    
    \For{$r = 1$ \textbf{to} $I_{max}$}
        \State $y_r \leftarrow \text{LLM}(M)$
    
        \If{$y_r$ is a valid \texttt{tool\_call}}
            \State $Q_r \leftarrow \operatorname{ParseAndValidate}(y_r)$
            \State $M \leftarrow \operatorname{Append}(M,y_r)$
            \State $\mathcal{R}_r \leftarrow \operatorname{Retrieve}_K(\mathcal{DB}_u^{\mathrm{lib}},Q_r)$
            \State $\mathcal{R} \leftarrow \mathcal{R} \cup \mathcal{R}_r$
            \State $M \leftarrow \operatorname{Append}(M,\texttt{tool\_response}:\mathcal{R}_r)$
    
        \ElsIf{$y_r$ is \texttt{answer}}
            \State $y_r \leftarrow \operatorname{PrivateRepair}(M,C,\mathcal{R},y_r;B)$
            \If{$y_r \neq \textsc{Failure}$ and $\operatorname{Valid}(y_r,\ell_{gt})$}
                \State $M \leftarrow \operatorname{Append}(M,y_r)$
                \State \Return $M$
            \EndIf
        \EndIf
    \EndFor
    \State $y_f \leftarrow \text{LLM}(M \mid \ell_{gt})$
    \State $y_f \leftarrow \operatorname{PrivateRepair}(M,C,\mathcal{R},y_f,\ell_{gt};B)$
    \If{$y_f \neq \textsc{Failure}$ and $\operatorname{Valid}(y_f,\ell_{gt})$}
        \State $M \leftarrow \operatorname{Append}(M,y_f)$
        \State \Return $M$
    \Else
        \State \Return \textsc{Discard}
    \EndIf
    \end{algorithmic}
    \end{algorithm}

Algorithm~\ref{alg:contrastive_multiturn} constructs each supervised trajectory as an ordered message history $M$. The history begins with a system prompt $msg_{sys}$ containing the task instruction, task ontology, and tool and answer schemas, followed by a user message $msg_{user}(C)$ containing the current context. We allow at most $I_{max}=3$ trajectory turns and up to $B=6$ private repair attempts for each answer candidate. At each trajectory turn, the teacher either produces a structured answer or issues a retrieval call containing one or more candidate intent explanations $Q_r$, with the number of candidates determined from the current context. For a retrieval call, the corresponding same-user records $\mathcal{R}_r$ are retrieved from $\mathcal{DB}_u^{\mathrm{lib}}$, added to the cumulative set $\mathcal{R}$, and recorded with the tool call and response in $M$. Before an answer candidate can be appended to $M$, $\operatorname{PrivateRepair}$ checks whether its reasoning and intent explanation support the predicted intent and are grounded in $C$ and $\mathcal{R}$, revising the candidate for up to $B$ attempts when needed. Candidates produced during the trajectory turns undergo private repair before label validation against $\ell_{gt}$ and format validation. A repaired candidate is appended to $M$ only if it passes these validations; otherwise, generation proceeds to the next trajectory turn. If no answer has been accepted after $I_{max}$ trajectory turns, the teacher uses $\ell_{gt}$ to generate and privately repair one final candidate. The ground-truth conditioning is confined to teacher-side target construction, and only a candidate that passes label and format validation is appended to $M$. The disjoint library and trajectory splits prevent the target instance from being retrieved.

\section{IntPro: Intent Inference with Selective Retrieval}
\label{sec:method}

IntPro learns direct and retrieval-conditioned inference through a two-stage training procedure. SFT learns both paths from validated teacher trajectories. GRPO then samples new trajectories and jointly optimizes intent prediction and the retrieval decision, with retrieval-conditioned rollouts using the per-user intent library defined in Section~\ref{sec:data}. Figure~\ref{training-pipeline} illustrates this training procedure.

\subsection{Supervised Fine-Tuning}
The first stage fine-tunes the proxy agent on the validated trajectories from Algorithm~\ref{alg:contrastive_multiturn}. Let $x$ denote the token sequence of a complete multi-turn trajectory and $\mathcal{T}_{\mathrm{asst}}$ the positions belonging to assistant messages. Assistant-only masking minimizes
\begin{equation}
\mathcal{L}_{\mathrm{SFT}}(\theta)
=-\frac{1}{|\mathcal{T}_{\mathrm{asst}}|}
\sum_{j\in\mathcal{T}_{\mathrm{asst}}}
\log \pi_\theta(x_j\mid x_{<j}),
\label{eq:sft_loss}
\end{equation}
where $\pi_\theta$ is the student policy. Assistant-only masking applies the loss to all assistant tokens retained in each teacher trajectory. The remaining messages, including the system prompt, user context, and tool responses, serve only as conditioning context. Direct trajectories supervise the structured intent output from $C$, whereas retrieval-conditioned trajectories additionally supervise candidate intent explanation generation, the retrieval call, and final intent inference conditioned on the returned historical records.

\subsection{GRPO with Retrieval-Decision Reward}
\label{sec:tool_reward}

Building on the two inference behaviors established by SFT, GRPO \cite{shao2024deepseekmath} refines intent prediction and the retrieval decision using trajectory-level rewards. For each context $C$, the policy samples a group of complete trajectories $\{o_i\}_{i=1}^{G}$ and computes centered advantages:
\begin{equation}
\bar R=\frac{1}{G}\sum_{j=1}^{G}R(o_j),\qquad
\hat A_i=R(o_i)-\bar R,
\label{eq:grpo_advantage}
\end{equation}

Let $\mathcal{T}_i$ be the policy-generated completion-token positions included in rollout $o_i$, let $L$ be the configured maximum completion length, and define
\begin{equation}
\rho_{i,t}(\theta)=
\frac{\pi_\theta(o_t^i\mid C,o_{<t}^i)}
     {\pi_{\theta_{\mathrm{old}}}(o_t^i\mid C,o_{<t}^i)}.
\end{equation}
With $s_\epsilon(\rho,A)=\min\{\rho A,\operatorname{clip}(\rho,1-\epsilon,1+\epsilon)A\}$, we minimize
\begin{equation}
\mathcal{L}_{\text{GRPO}}(\theta)
=-\frac{1}{LG}\sum_{i=1}^{G}
\sum_{t\in\mathcal{T}_i}
s_\epsilon(\rho_{i,t},\hat A_i),
\label{eq:grpo_loss}
\end{equation}
where the importance-ratio denominator is computed under the old policy.

To optimize the retrieval decision together with intent prediction, we define a trajectory-level reward for each rollout. Let $R_{\mathrm{acc}}(o_i)=\mathbbm{1}[\ell_{pred}^{(i)}=\ell_{gt}]$. The reward combines output validity and final intent correctness with a retrieval-decision term:
\begin{equation}
R(o_i)=R_{\mathrm{format}}(o_i)+R_{\mathrm{acc}}(o_i)
+\lambda_{\mathrm{dec}}R_{\mathrm{decision}}(o_i).
\label{eq:total_reward}
\end{equation}
$R_{\mathrm{format}}(o_i)=1$ when the rollout follows the required answer and intent explanation schemas and, when retrieval is invoked, the tool-call schema; otherwise, $R_{\mathrm{format}}(o_i)=0$. During GRPO, each retrieval-conditioned rollout contains one retrieval call with a nonempty set of candidate intent explanations whose size is determined by the policy. We set $\lambda_{\mathrm{dec}}=1$ for IntPro and $\lambda_{\mathrm{dec}}=0$ for Naive GRPO, so the latter removes the retrieval-decision term while retaining the format and intent-accuracy rewards.
Both inference paths are evaluated by the final prediction: $R_{\mathrm{decision}}$ conditions on the selected path and on $R_{\mathrm{acc}}(o_i)$, so a retrieval rollout earns credit only when the retrieval-conditioned prediction is correct. The group success rate
\begin{equation}
\alpha_\mathcal{G}=\frac{1}{G}\sum_{i=1}^{G}R_{\mathrm{acc}}(o_i)
\end{equation}
measures how often the current policy predicts the correct intent across the $G$ rollouts for the same context. Let $z_i\in\{0,1\}$ denote the inference path selected in rollout $o_i$, where $z_i=0$ corresponds to direct inference and $z_i=1$ indicates that the rollout invokes retrieval. We use $\alpha_\mathcal{G}$ as a graded weight for direct inference and $1-\alpha_\mathcal{G}$ as the complementary weight for retrieval-conditioned inference:
\begin{equation}
R_{\mathrm{decision}}(o_i)
=
\begin{cases}
\alpha_\mathcal{G},
& z_i=0,\ R_{\mathrm{acc}}(o_i)=1,\\
-(1-\alpha_\mathcal{G}),
& z_i=0,\ R_{\mathrm{acc}}(o_i)=0,\\
1-\alpha_\mathcal{G},
& z_i=1,\ R_{\mathrm{acc}}(o_i)=1,\\
-\alpha_\mathcal{G},
& z_i=1,\ R_{\mathrm{acc}}(o_i)=0.
\end{cases}
\label{eq:retrieval_decision_reward}
\end{equation}

The four cases use the same group statistic to assign complementary weights to the direct and retrieval-conditioned outcomes. When group success is high, a correct direct prediction receives more credit and an incorrect retrieval-conditioned prediction incurs a stronger penalty. When group success is low, a correct retrieval-conditioned prediction receives more credit and an incorrect direct prediction incurs a stronger penalty. The weights vary continuously with $\alpha_\mathcal{G}$.

This weighting admits a direct characterization in expectation. For a given context, let $p_{\mathrm{dir}}$ and $p_{\mathrm{ret}}$ denote the probabilities that the current policy predicts the correct intent through direct and retrieval-conditioned inference, respectively, and treat $\alpha_\mathcal{G}$ as the observed group statistic. The expected decision reward is $p_{\mathrm{dir}}+\alpha_\mathcal{G}-1$ for a direct rollout and $p_{\mathrm{ret}}-\alpha_\mathcal{G}$ for a retrieval rollout, so with $\lambda_{\mathrm{dec}}=1$ the combined accuracy and decision terms favor the retrieval path exactly when
\begin{equation}
p_{\mathrm{ret}}-p_{\mathrm{dir}} > \alpha_\mathcal{G}-\tfrac{1}{2}.
\label{eq:retrieval_condition}
\end{equation}
Equation~\ref{eq:retrieval_condition} shows how group success changes the accuracy difference required for retrieval to receive higher reward. When $\alpha_{\mathcal G}$ is high, retrieval must provide a larger advantage over direct inference. When $\alpha_{\mathcal G}$ is low, the threshold becomes negative, creating a bounded preference for retrieval on contexts that the current policy rarely solves. Because $\alpha_{\mathcal G}\in[0,1]$, the threshold remains within $-\frac{1}{2},\frac{1}{2}$, while the rewards for both paths continue to depend on final intent correctness.

\section{Experimental Evaluation}
\label{sec:experiments}

Our experiments first evaluate IntPro across scenarios and model backbones and characterize the Direct-and-Retrieval SFT initialization used before GRPO. We then test whether intent explanations support intent-aligned retrieval and whether the learned retrieval decision improves over fixed policies. Retrieval-budget and class-level analyses examine selected sources of variation, and paired cases illustrate how retrieved history can either help or hurt an intent prediction.

\subsection{Experimental Setup}

\subsubsection{Data and Settings}
We conduct experiments across three scenarios: dialogue intent understanding (MIntRec2.0), social-media intent understanding (Weibo Post-Sync), and reading intent understanding (Highlight-Intent). We instantiate the context and intent-library variables from Sections~\ref{sec:context_intent} and~\ref{sec:history_library} as follows.

\textbf{MIntRec2.0.} MIntRec2.0 \cite{zhang2024mintrec} is a benchmark for multimodal intent recognition in multi-party dialogues, comprising thirty fine-grained intent labels spanning two semantic super-categories (express emotion/attitude vs. achieve goal). We pool the released train, development, and test sources and remove targets labeled \texttt{UNK} or attributed to a placeholder program name, leaving 8,383 target utterances in 1,234 dialogues. Complete dialogues, rather than individual utterances, are assigned to the library, training, and test partitions, so all target utterances from a dialogue remain together. We use the preceding dialogue turns as the interaction environment $E$, the current utterance as action $A$, and the scripted-character identifier to associate interactions with the same person.

\textbf{Weibo Post-Sync.} The Weibo Post-Sync dataset is derived from a study of cross-platform posting behavior on Chinese social platforms \cite{10.1145/3457986}, whose anonymized posts are annotated with one of seven posting intents, together with content-category and topic labels. From this annotated corpus we select users who each have a sufficient posting volume and cover multiple posting-intent categories, and retain their posts with complete identifier, content, and intent fields, yielding 3,184 posts with unique post identifiers. Each post is assigned to one partition. We use the platform, content category, and topic hashtags as the interaction environment $E$, the current post as $A$, and the anonymized user ID to associate interactions with the same person.

\textbf{Highlight-Intent.} Highlight-Intent is an English reading-assistance dataset for context-aware intent inference. Each record pairs a \emph{stimulus}, the article and surrounding passage the reader was reading, with the span the reader highlighted and the reading intent behind that highlight. Intents follow a 12-class taxonomy covering term-directed, data-directed, and opinion-directed information needs. The complete dataset contains 3,297 records from a highlighting-and-annotation study in which 23 readers read 19 English articles. All readers participated with informed consent to the collection and research use of their highlighting and annotation records. Records from the same deduplicated reader--stimulus group remain in one partition. We use the article and surrounding passage as the interaction environment $E$, the highlighted span as the action $A$, and the reader identifier to associate interactions with the same person.

We divide each dataset into non-overlapping intent-library, training, and test sets. Records belonging to the same dialogue or user--stimulus group remain in the same set to prevent information from crossing partitions. The assignment targets an approximate 30\%/50\%/20\% ratio and, for sparse user--intent groups, places records in the intent library and training set before creating test records. The intent library then provides the fixed source of same-user history, while the separate training and test sets are used for proxy learning and evaluation. Table~\ref{tab:experimental_protocol} reports the resulting sizes, and all evaluations use the complete test sets.

\begin{table}[t]
\centering
\caption{Dataset partitions after data cleaning.}
\label{tab:experimental_protocol}
\scriptsize
\begin{tabular}{lrrrr}
\toprule
\textbf{Dataset} & \textbf{Total} & \textbf{Intent library} & \textbf{Training set} & \textbf{Test set} \\
\midrule
MIntRec2.0 & 8,383 & 2,340 & 4,363 & 1,680 \\
Weibo Post-Sync & 3,184 & 775 & 1,923 & 486 \\
Highlight-Intent & 3,297 & 980 & 1,660 & 657 \\
\bottomrule
\end{tabular}
\end{table}

\subsubsection{Models and Baselines}
We evaluate IntPro using Qwen2.5-3B, Llama3.2-3B, and Qwen3-4B as model backbones. The main controlled comparison is between Naive GRPO and IntPro. Both start from the same Direct-and-Retrieval SFT checkpoint and share the GRPO training setup and base reward components. Naive GRPO sets $\lambda_{\mathrm{dec}}=0$, whereas IntPro sets $\lambda_{\mathrm{dec}}=1$. Two SFT conditions locate this comparison relative to supervised training. Answer-only SFT learns direct intent-label prediction without retrieval trajectories. Direct-and-Retrieval SFT learns both inference paths and is evaluated before GRPO with the SFT model choosing between them. We also report Gemini-3-flash and Qwen3-30B-A3B as model-scale references. Their Direct condition provides the current context, task ontology, and answer schema without retrieved records. Their Retrieval-Augmented condition invokes retrieval for every test instance. The same model first generates one candidate intent explanation from the current context, after which BGE-M3 retrieves from the same per-user intent library using the procedure in Section~\ref{sec:data}. We use $K=4$ for Highlight-Intent and MIntRec2.0 and $K=2$ for Weibo Post-Sync, and the same model predicts the structured intent from the current context and returned records. We additionally report the discriminative BERT-base and RoBERTa-base models.

All SFT variants use AdamW, a maximum sequence length of 4,096 tokens, three training epochs, and a learning rate of $1\times10^{-5}$. The GRPO variants use AdamW for one epoch with a learning rate of $1\times10^{-6}$. We use the Dr.\ GRPO loss \cite{liu2025understanding} with $G=8$, $\epsilon=0.2$, no reward-standard-deviation scaling, and no reference-policy KL term. For generative inference, we use a temperature of 0.7, top-$p$ of 0.95. The student models are trained using two NVIDIA A100 GPUs.

\subsubsection{Metrics}
The main comparisons report Accuracy (Acc), Macro-F1 (M-F1), and Weighted-F1 (W-F1). Accuracy is the proportion of correct intent predictions, and Accuracy differences are computed from test-set correct counts before rounding. Macro-F1 averages the class-wise F1 scores equally, while Weighted-F1 weights them by class support. The training diagnostics additionally report Pass@4, which records whether at least one of four sampled predictions is correct, and the Generalization Gap between training and test Accuracy. For the radar-chart visualization, we invert the min-max normalized Generalization Gap within each dataset, so a higher plotted value denotes a smaller gap.

\subsection{Overall Intent Inference Performance}

We first evaluate IntPro's final intent-inference performance across datasets and model backbones. The controlled comparison is IntPro versus Naive GRPO because both start from the same Direct-and-Retrieval SFT checkpoint, share the GRPO training setup and base reward components, and differ only in the retrieval-decision reward. The SFT conditions and external models provide supervised-training and model-scale references.

\begin{table*}[htbp]
    \centering
    \footnotesize
    \setlength{\tabcolsep}{1mm}
    \caption{Overall context-aware intent-inference performance across three datasets. IntPro and Naive GRPO share the Direct-and-Retrieval SFT initialization; bold marks the best result within each backbone block.}
    \begin{tabular}{l ccc ccc ccc}
    \toprule
    \multirow{2}{*}{\textbf{Model}} & \multicolumn{3}{c}{\textbf{Highlight-Intent}} & \multicolumn{3}{c}{\textbf{MIntRec2.0}} & \multicolumn{3}{c}{\textbf{Weibo Post-Sync}} \\
    & Acc (\%) & M-F1 (\%) & W-F1 (\%) & Acc (\%) & M-F1 (\%) & W-F1 (\%) & Acc (\%) & M-F1 (\%) & W-F1 (\%) \\
    \midrule
    \multicolumn{10}{c}{\textit{LLM and Discriminative Baselines}} \\
    \midrule
    Gemini-3-flash (Direct)& 32.42& 16.27& 29.35& 41.43& 29.68& 43.19& 46.30& 41.56& 47.63\\
    Gemini-3-flash (Retrieval-Augmented)& 40.18& 18.70& 35.42& 46.13& 31.96& 47.12& 67.70& 44.49& 62.08\\
    Qwen3-30B-A3B (Direct)& 29.22& 13.82& 26.33& 38.93& 26.06& 39.59& 42.80& 32.15& 37.88\\
    Qwen3-30B-A3B (Retrieval-Augmented)& 36.53& 17.50& 31.14& 41.96& 30.83& 43.46& 63.99& 40.06& 57.27\\
    BERT-base& 38.36& 13.64& 30.07& 42.14& 23.71& 32.94& 67.49& 33.09& 60.80\\
    RoBERTa-base& 45.05& 23.07& 37.95& 46.25& 28.53 & 41.91& 63.37& 31.97& 57.50\\
    \midrule
    \multicolumn{10}{c}{\textit{Qwen2.5-3B}} \\
    \midrule
    Answer-only SFT& 44.90& 23.85& 37.40& 63.93& 58.47 & 63.02& 70.37& 52.86& 69.19\\
    Direct-and-Retrieval SFT& 42.47& 22.15& 35.89& 60.54& 56.04& 61.26& 66.87& 51.74& 66.13\\
    \addlinespace
    Naive GRPO& 43.23& 26.79& 38.43& 62.44& 57.65& 62.14& 68.52& 55.72& 69.83\\
    \rowcolor{drssbg}
    IntPro (Ours)& \textbf{47.03}& \textbf{29.32}& \textbf{41.67}& \textbf{65.83}& \textbf{58.73} & \textbf{64.26}& \textbf{75.10}& \textbf{61.69}& \textbf{74.85}\\
    \midrule
    \multicolumn{10}{c}{\textit{Llama3.2-3B}} \\
    \midrule
    Answer-only SFT& 36.53& 24.51& 35.93& 57.74& 48.59& 56.62& 66.87& 51.60& 66.28\\
    Direct-and-Retrieval SFT& 34.55& 23.68& 33.97& 55.36& 47.24& 54.87& 64.40& 50.09& 63.66\\
    \addlinespace
    Naive GRPO& 43.07& 27.16& 39.25& 57.92& 43.71& 56.97& 67.49& 51.57& 67.41\\
    \rowcolor{drssbg}
    IntPro (Ours)& \textbf{45.66}& \textbf{28.93}& \textbf{40.84}& \textbf{59.58}& \textbf{49.17}& \textbf{58.29}& \textbf{70.58}& \textbf{55.53}& \textbf{69.48}\\
    \midrule
    \multicolumn{10}{c}{\textit{Qwen3-4B}} \\
    \midrule
    Answer-only SFT& 48.40& 30.71& 42.89& 62.74 & 52.94 & 63.22& 77.16& 62.35 & 76.81\\
    Direct-and-Retrieval SFT& 44.75& 27.83& 39.35& 63.45& 54.54& 64.02& 72.84& 58.63& 72.41\\
    \addlinespace
    Naive GRPO& 47.79& 28.93& 41.94& 66.85& 55.63& 67.22& 76.54& 60.71& 76.19\\
    \rowcolor{drssbg}
    IntPro (Ours)& \textbf{51.45}& \textbf{33.07}& \textbf{45.93}& \textbf{69.70}& \textbf{59.24}& \textbf{68.95}& \textbf{79.63}& \textbf{64.83}& \textbf{78.90}\\
    \bottomrule
    \end{tabular}

    \label{tab:performance_GRPO}
\end{table*}

IntPro achieves the highest Accuracy, Macro-F1, and Weighted-F1 for every model backbone in Table~\ref{tab:performance_GRPO}. Relative to Naive GRPO, adding the retrieval-decision reward improves all 27 combinations of dataset, model backbone, and metric. The gains range from 1.67 to 6.58 percentage points in Accuracy, from 1.08 to 5.97 points in Macro-F1, and from 1.32 to 5.02 points in Weighted-F1. This consistency across datasets and model backbones supports the retrieval-decision reward beyond a gain confined to one setting.

Answer-only SFT has higher Accuracy than Direct-and-Retrieval SFT in eight of the nine combinations of dataset and model backbone; the only reversal is a 0.71-point advantage for Direct-and-Retrieval SFT on MIntRec2.0 with Qwen3-4B. After GRPO, IntPro exceeds both SFT variants on every reported metric. These rows characterize the supervised starting point: Direct-and-Retrieval SFT supplies the two inference paths used by both GRPO variants, while the IntPro--Naive GRPO comparison tests the retrieval-decision reward under this shared initialization.

The external references provide model-scale context for these gains. Retrieval-Augmented Gemini-3-flash and Qwen3-30B-A3B improve over their Direct counterparts on every metric, with the largest differences on Weibo Post-Sync. IntPro with each model backbone exceeds all four LLM reference conditions and both discriminative baselines on all reported metrics across the three datasets.

\subsubsection{SFT Initialization Strategies}
\begin{figure}[t]
\centering
\includegraphics[width=0.5\textwidth]{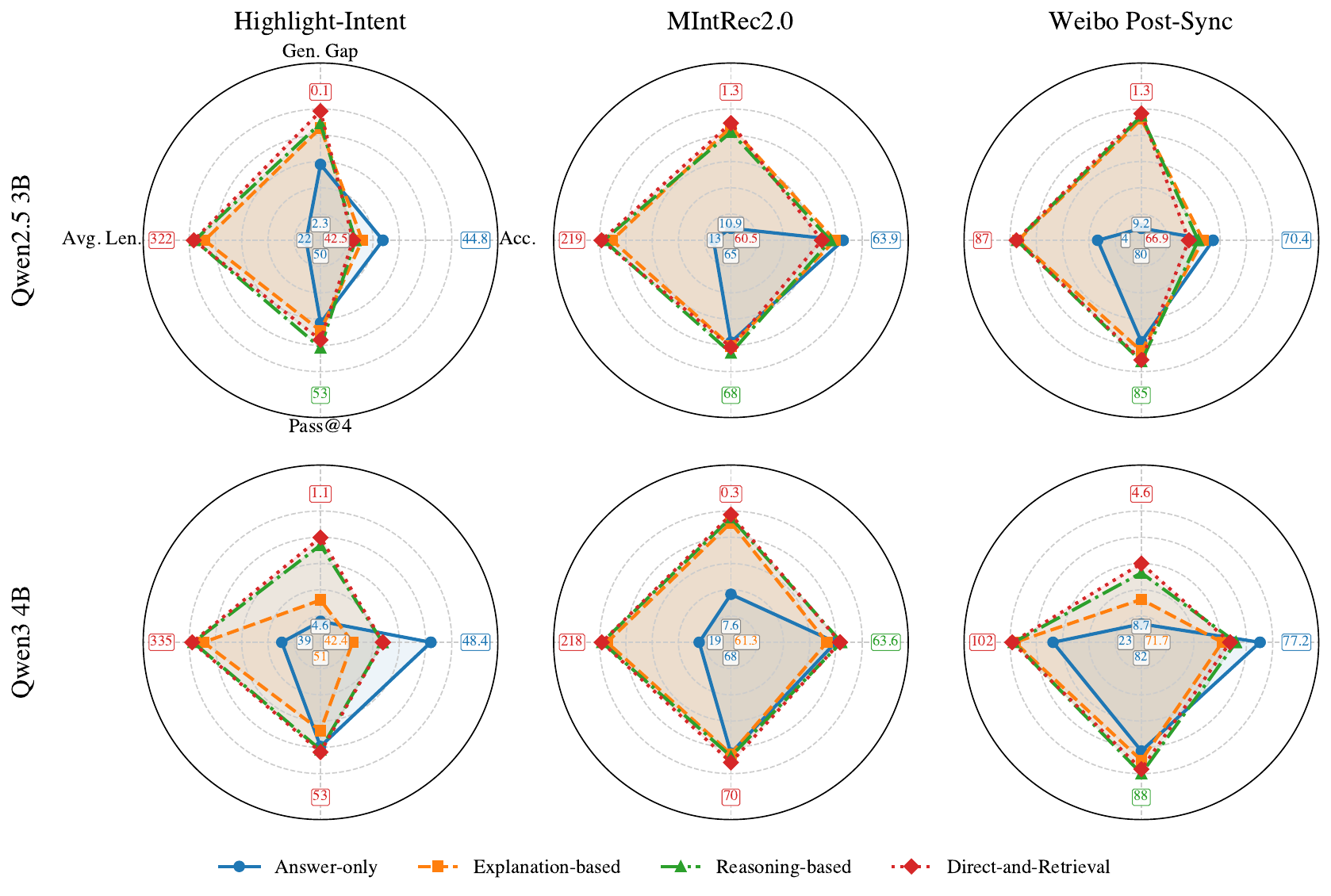}
\caption{Four supervised fine-tuning formats across two models and three datasets. The Generalization Gap axis is inverted for radial scaling.}
\label{compare-sft}
\end{figure}

Table~\ref{tab:performance_GRPO} shows that Direct-and-Retrieval SFT is usually weaker than Answer-only SFT in single-shot Accuracy, although it establishes the two inference paths required by GRPO. Figure~\ref{compare-sft} characterizes this supervised initialization through four SFT formats: \textbf{Answer-only} outputs only the intent label; \textbf{Explanation-based} also outputs an intent explanation; \textbf{Reasoning-based} adds an explicit reasoning trace; and \textbf{Direct-and-Retrieval} combines direct trajectories with retrieval trajectories containing a tool call, returned historical records, and the final structured intent output. All four formats are trained for three epochs.

The four formats show different supervised-training profiles before GRPO. Answer-only SFT has the highest training Accuracy and the largest train--test gap. Direct-and-Retrieval SFT has lower single-shot Accuracy than Answer-only SFT in five of the six panels, yet it has the smallest Generalization Gap in every panel. Its longer outputs reflect the tool calls and returned historical records in the retrieval trajectories. Direct-and-Retrieval SFT has the highest Pass@4 in two Qwen3-4B panels, while Reasoning-based SFT has the highest Pass@4 in the other four. The checkpoint used in the main GRPO comparison is therefore motivated by its dual-path trajectories rather than by being the strongest stand-alone classifier.

\begin{table*}[t]
    \centering
    \caption{Comparison of retrieval representations for annotated historical records across datasets.}
    \label{tab:explanation_discriminability}
    \scriptsize
    \begin{tabular}{lp{4.0cm}cccc}
    \toprule
    \textbf{Dataset} & \textbf{Input Form} & \textbf{Intra-Sim}$\uparrow$ & \textbf{Inter-Sim}$\downarrow$ & \textbf{User LOO Acc.}$\uparrow$ & \textbf{Global R@1}$\uparrow$ \\
    \midrule
    \multirow{4}{*}{MIntRec2.0}
    & Action Only & 0.5218 & \textbf{0.4843} & 0.2295 & 0.3171 \\
    & Full Context & 0.5306 & 0.5239 & 0.0848 & 0.0940 \\
    & Free-form Context-to-Intent Description & 0.5979 & 0.5584 & 0.3386 & 0.4376 \\
    % \rowcolor{drssbg}
    & Intent explanation & \textbf{0.6067} & 0.5598 & \textbf{0.3818} & \textbf{0.5051} \\
    \midrule
    \multirow{4}{*}{Weibo Post-Sync}
    & Action Only & 0.4633 & \textbf{0.4045} & 0.3690 & 0.6438 \\
    & Full Context & 0.5581 & 0.5173 & 0.3781 & 0.6215 \\
    & Free-form Context-to-Intent Description & 0.6107 & 0.5871 & 0.4032 & 0.7286 \\
    % \rowcolor{drssbg}
    & Intent explanation & \textbf{0.6428} & 0.5376 & \textbf{0.4533} & \textbf{0.8043} \\
    \midrule
    \multirow{4}{*}{Highlight-Intent}
    & Action Only & 0.3497 & \textbf{0.3379} & 0.2439 & 0.3112 \\
    & Full Context & 0.3980 & 0.3808 & 0.2429 & 0.2927 \\
    & Free-form Context-to-Intent Description & 0.5277 & 0.5521 & 0.3571 & 0.4592 \\
    % \rowcolor{drssbg}
    & Intent explanation & \textbf{0.5910} & 0.4871 & \textbf{0.3582} & \textbf{0.4662} \\
    \bottomrule
    \end{tabular}
\end{table*}

\subsection{Intent-Aligned Retrieval with Intent Explanations}
The value of retrieval-conditioned inference depends on whether its query and library representation surface historical records related to the underlying intent. We therefore test the two retrieval roles of intent explanations: representing annotated historical records in the per-user intent library and forming queries from the current context. We test these roles separately because historical explanations are constructed from annotated records $(C,\ell)$, while test-time queries are formed from $C$. Same-intent retrieval is the operational measure of intent alignment in both analyses.

\subsubsection{Intent Explanation as a Retrieval Representation}
\label{sec:discriminability}
The library-side question is whether intent explanations make annotated historical records more retrievable by intent than their observable action, full context, or a free-form context-to-intent description. We embed each input form using BGE-M3 \cite{chen2024m3}. Intra-class and Inter-class Similarity measure the average cosine similarity within and across intent categories. \textbf{User LOO Acc} uses each record in turn to retrieve from the same user's remaining records and macro-averages the resulting accuracy across users. \textbf{Global R@1} is the proportion of records whose nearest neighbor in the remaining dataset-wide pool has the same intent label. Here, ``user'' denotes a scripted character in MIntRec2.0, an anonymized account in Weibo Post-Sync, and a reader in Highlight-Intent.

We compare four input forms. \textbf{Action Only} uses the dataset-specific observable action $A$ (the current utterance, post, or highlighted span) without the interaction environment $E$, while \textbf{Full Context} uses $C=(E,A)$. \textbf{Free-form Context-to-Intent Description}, $p_{\mathrm{free}}=\operatorname{LLM}_{\mathrm{free}}(C,\ell)$, lets the LLM describe how the context expresses the annotated intent in one sentence (e.g., ``By highlighting how biochemical self-incompatibility lets homomorphic flowers distinguish self from non-self pollen, the user expresses a Define or Explain intent toward the mechanism''). The \textbf{intent explanation} defined in Section~\ref{sec:context_intent} retains the original action, abstracts its functional goal, and includes a cue-to-intent rationale. Both generated forms are constructed independently for each labeled record; neither generator receives other same-user records.

\begin{figure}[t]
\centering
\includegraphics[width=0.5\textwidth]{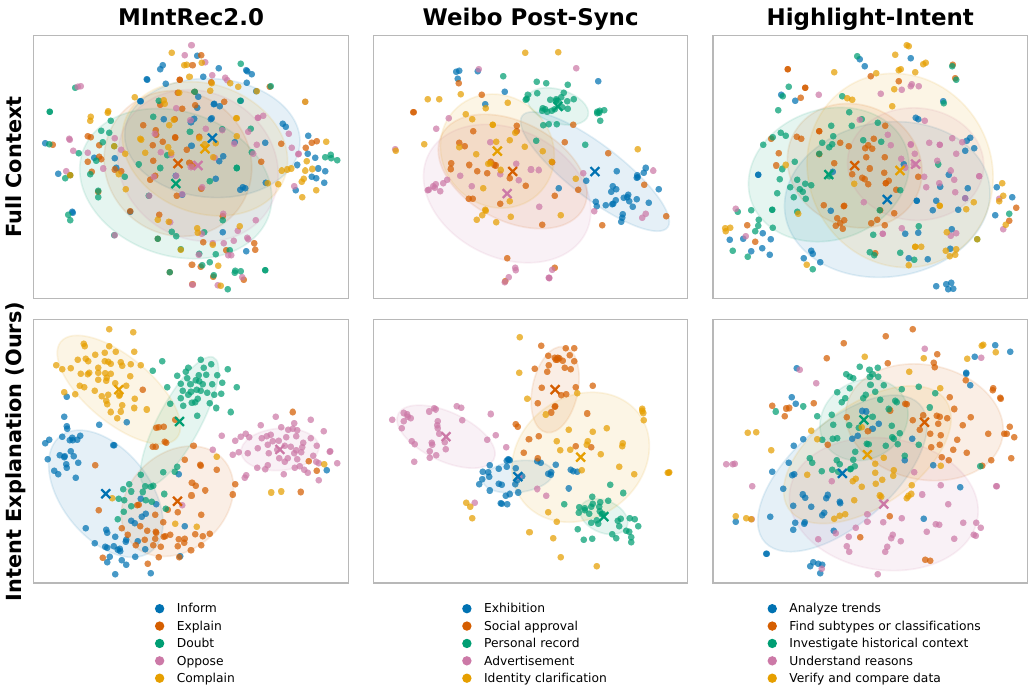}
\caption{t-SNE projections of Full Context (top) and the intent explanation (bottom) on MIntRec2.0, Weibo Post-Sync, and Highlight-Intent. Within each dataset, the two rows use the same sampled records; each panel is projected independently and is shown descriptively.}
\label{fig:tsne_explanation}
\end{figure}

The intent explanation provides the strongest library-side retrieval representation in Table~\ref{tab:explanation_discriminability}: it has the highest Intra-Sim, User LOO Acc, and Global R@1 on all three datasets. Relative to Full Context, its Global R@1 increases by 17.35--41.11 percentage points, supporting the complete intent explanation as a more intent-aligned library-side retrieval representation under the evaluated setting. Its Intra-Sim exceeds its Inter-Sim by 0.047, 0.105, and 0.104 on MIntRec2.0, Weibo Post-Sync, and Highlight-Intent, respectively, the largest separation among the four input forms in each dataset. The advantage over the Free-form Context-to-Intent Description is clear on MIntRec2.0 and Weibo Post-Sync and marginal on Highlight-Intent, where User LOO Acc and Global R@1 increase by 0.0011 and 0.0070. Figure~\ref{fig:tsne_explanation} provides a descriptive comparison with Full Context. The intent explanation projections show clearer local grouping across all three datasets, with less pronounced separation on Highlight-Intent than on MIntRec2.0 and Weibo Post-Sync.

\subsubsection{Intent Explanation as a Retrieval Query}
\label{sec:retrieval_effect}
The test-time question is whether an intent explanation generated from the current context retrieves more intent-aligned historical records and improves the final intent prediction. We use Gemini-3-flash as a common inference model on the complete test partitions in Table~\ref{tab:experimental_protocol}. In the intent explanation condition, it generates one candidate intent explanation from $C$, which BGE-M3 embeds as the retrieval query. Action Only and Full Context instead use $A$ or $C$ as the query, and Direct gives the model only $C$ and the task ontology. Each retrieval condition uses one query per test instance, after which the same inference model predicts from the current context and the returned records. Same-intent Hit@$K$ records whether the Top-$K$ results contain at least one record with the annotated intent. MIntRec2.0 and Highlight-Intent use $K=4$, and Weibo Post-Sync uses $K=2$, following the training-set selection.

\begin{table}[t]
    \centering
    \scriptsize
    \caption{Evaluating intent explanations as retrieval queries for intent inference.}
    \label{tab:retrieval_reader}

    \textbf{(a) Same-intent Hit@$K$ (\%)}\par
    \begin{tabular}{lccc}
    \toprule
    \textbf{Condition} & \textbf{MIntRec2.0} & \shortstack{\textbf{Weibo}\\\textbf{Post-Sync}} & \shortstack{\textbf{Highlight-}\\\textbf{Intent}} \\
    \midrule
    Action Only & 52.86 & 78.19 & 56.32 \\
    Full Context & 27.14 & 76.95 & 56.62 \\
    Intent explanation & \textbf{56.43} & \textbf{80.45} & \textbf{58.90} \\
    \bottomrule
    \end{tabular}

    \vspace{0.5em}
    \textbf{(b) Intent Accuracy (\%)}\par
    \begin{tabular}{lccc}
    \toprule
    \textbf{Condition} & \textbf{MIntRec2.0} & \shortstack{\textbf{Weibo}\\\textbf{Post-Sync}} & \shortstack{\textbf{Highlight-}\\\textbf{Intent}} \\
    \midrule
    Direct & 49.70 & 47.12 & 35.92 \\
    Action Only & 55.06 & 59.67 & 36.99 \\
    Full Context & 49.76 & 58.44 & 37.60 \\
    Intent explanation & \textbf{55.60} & \textbf{63.99} & \textbf{38.05} \\
    \bottomrule
    \end{tabular}
\end{table}

The intent explanation achieves the highest Same-intent Hit@$K$ and Intent Accuracy on all three datasets in Table~\ref{tab:retrieval_reader}. Compared with the stronger raw-query form in each dataset, it improves Hit@$K$ by 2.26--3.57 percentage points and Accuracy by 0.46--4.32 points. Compared with Direct inference, the final Accuracy gain ranges from 2.13 points on Highlight-Intent to 16.87 points on Weibo Post-Sync. The two panels also show that better retrieval does not translate uniformly into an equal prediction gain: on MIntRec2.0, a 3.57-point Hit@$K$ increase over Action Only accompanies a 0.54-point Accuracy increase. Together, Tables~\ref{tab:explanation_discriminability} and~\ref{tab:retrieval_reader} support intent explanations as a shared intent-oriented natural-language form for representing annotated historical records and forming intent-aligned queries from the current context. For a controlled comparison of query forms, each retrieval condition uses one query per test instance, after which the same inference model predicts from the current context and the returned records.

\subsection{Selective Retrieval Analysis}

Intent-aligned historical records do not benefit every inference context. We therefore test whether the Qwen3-4B IntPro model benefits from deciding whether to retrieve for each context, and whether any gain depends on which contexts invoke retrieval.

\subsubsection{Retrieval Policy Comparison}

Table~\ref{tab:retrieval_policy} compares three inference-time retrieval policies using the same trained IntPro checkpoint and otherwise identical evaluation settings. \textbf{No Retrieval} forces direct inference, \textbf{Always Retrieval} forces retrieval-conditioned inference, and \textbf{Selective Retrieval} lets IntPro decide whether to invoke retrieval for each context. The retrieval-conditioned paths use the budget selected on the training set, with $K=4$ for Highlight-Intent and MIntRec2.0 and $K=2$ for Weibo Post-Sync.

\begin{table}[t]
\centering
\caption{Context-aware intent-inference performance under fixed and selective retrieval policies using Qwen3-4B.}
\label{tab:retrieval_policy}
% \scriptsize
% \setlength{\tabcolsep}{3pt}
\begin{tabular}{llcc}
\toprule
\textbf{Policy} & \textbf{Dataset} & \textbf{Acc (\%)} & \textbf{M-F1 (\%)} \\
\midrule
\multirow{3}{*}{No Retrieval}
& Highlight-Intent & 48.71 & 31.29 \\
& MIntRec2.0 & 67.32 & 56.21 \\
& Weibo Post-Sync & 75.51 & 61.74 \\
\addlinespace
\multirow{3}{*}{Always Retrieval}
& Highlight-Intent & 45.36 & 28.04 \\
& MIntRec2.0 & 66.85 & 53.28 \\
& Weibo Post-Sync & 77.16 & 64.05 \\
\addlinespace
\multirow{3}{*}{\shortstack[l]{Selective Retrieval\\(IntPro)}}
& Highlight-Intent & 51.45 & 33.07 \\
& MIntRec2.0 & 69.70 & 59.24 \\
& Weibo Post-Sync & 79.63 & 64.83 \\
\bottomrule
\end{tabular}
\end{table}

Selective Retrieval achieves the highest Accuracy and Macro-F1 on all three datasets. Compared with the stronger fixed policy in each dataset, it improves Accuracy by 2.38--2.74 percentage points and Macro-F1 by 0.78--3.03 points. The stronger fixed policy itself changes with the dataset: No Retrieval is better on Highlight-Intent and MIntRec2.0, while Always Retrieval is better on Weibo Post-Sync. These results directly support deciding whether to invoke retrieval for each context instead of applying one fixed history-access policy to every prediction.

\subsubsection{Retrieval Decision Analysis}

The policy comparison leaves open whether Selective Retrieval benefits from its average retrieval rate or from the contexts for which it invokes retrieval. Table~\ref{tab:decision_anatomy} first reports the retrieval rate and the Accuracy achieved on the contexts assigned to each inference path. It then evaluates decision quality using paired direct and retrieval-conditioned outputs from the Qwen3-4B experiments. \textbf{Rate-matched random} matches Selective Retrieval's observed rate in expectation while randomly choosing which contexts take the retrieval-conditioned output. \textbf{Path oracle} selects the correct output whenever either path is correct and provides an upper bound over the same paired outputs.

\begin{figure}[t]
\centering
\includegraphics[width=0.95\columnwidth]{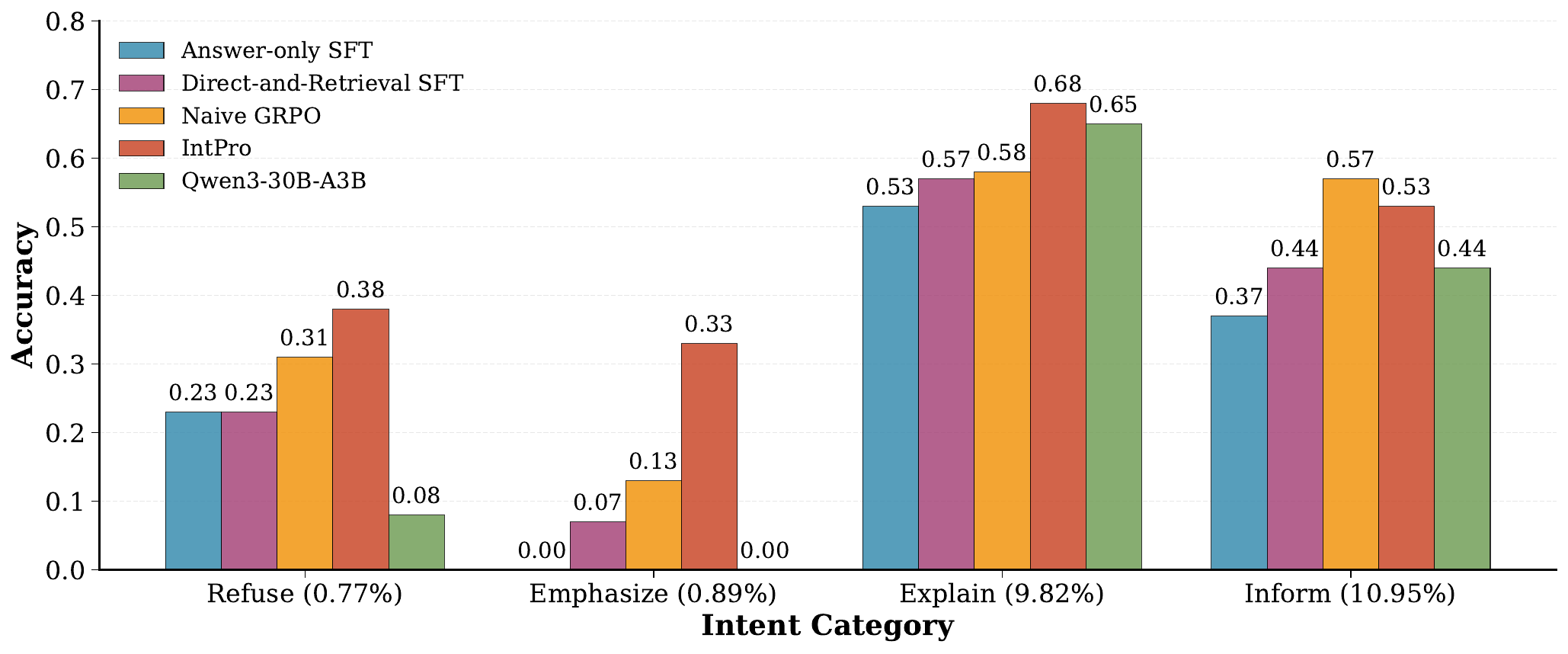}
\caption{Per-class Accuracy of the Qwen3-4B student variants and Qwen3-30B-A3B (Direct) on the two least-frequent and two most-frequent MIntRec2.0 intents.}
\label{histogram_class}
\end{figure}

\begin{table}[t]
\centering
\scriptsize
\caption{Retrieval frequency and decision quality for Selective Retrieval using Qwen3-4B.}
\label{tab:decision_anatomy}
\begin{tabular}{lccc}
\toprule
\textbf{Quantity} & \shortstack{\textbf{Highlight-}\\\textbf{Intent}} & \shortstack{\textbf{MIntRec}\\\textbf{2.0}} & \shortstack{\textbf{Weibo}\\\textbf{Post-Sync}} \\
\midrule
Retrieval rate (\%) & 12.02 & 19.82 & 45.06 \\
\addlinespace
Direct-inference Acc. (\%) & 51.38 & 69.93 & 79.03 \\
Retrieval-conditioned Acc. (\%) & 51.90 & 68.77 & 80.37 \\
\addlinespace
Rate-matched random (\%) & 48.30 & 67.23 & 76.26 \\
Selective Retrieval (\%) & 51.45 & 69.70 & 79.63 \\
Path oracle (\%) & 57.53 & 76.49 & 85.60 \\
\bottomrule
\end{tabular}
\end{table}

\begin{figure*}[!t]
\centering
\includegraphics[width=0.8\textwidth]{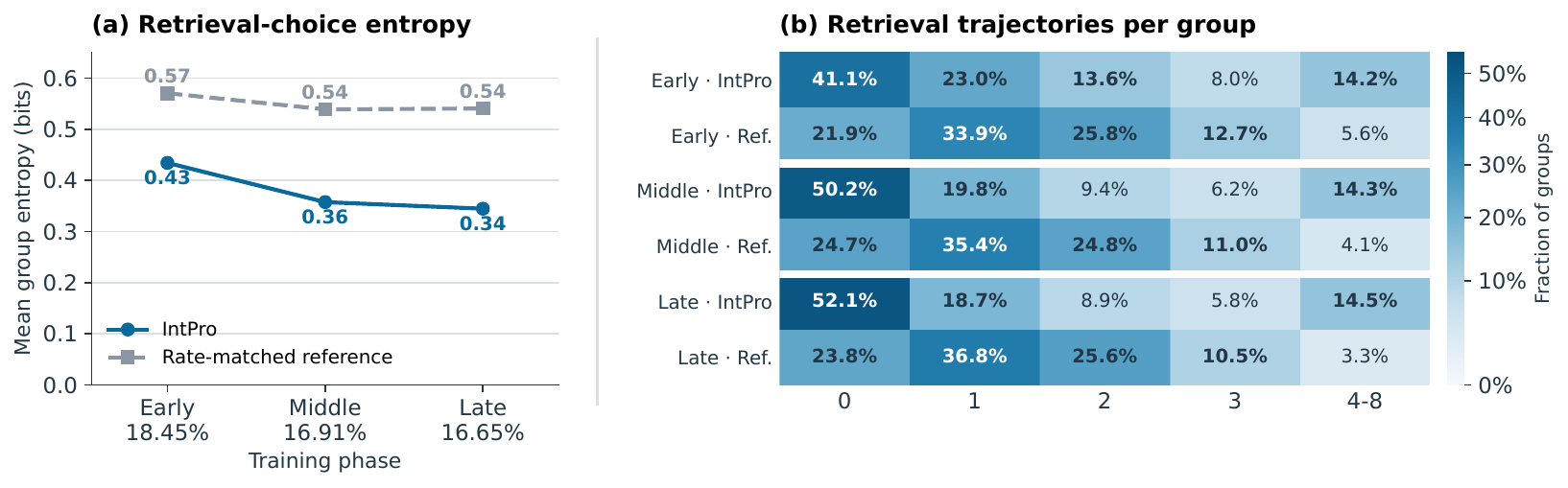}
\caption{Retrieval-choice dynamics of IntPro during GRPO training on MIntRec2.0. IntPro's observed group-level decisions are compared with a rate-matched independent-choice reference. (a) Mean retrieval-choice entropy. (b) Distribution of the number of retrieval trajectories in each group.}
\label{fig:grpo_decision_concentration}
\end{figure*}

Selective Retrieval exceeds the expected rate-matched random policy by 3.14, 2.48, and 3.37 percentage points in Accuracy on Highlight-Intent, MIntRec2.0, and Weibo Post-Sync, respectively. The random-policy score is computed from the same paired direct and retrieval-conditioned outputs using Selective Retrieval's observed retrieval rate, so this comparison controls for retrieval frequency and supports the value of choosing which contexts invoke retrieval. The path oracle remains 5.97--6.79 percentage points higher, quantifying the remaining headroom in path selection over the same paired outputs.

\paragraph{Retrieval behavior during IntPro training.}
To examine how IntPro's retrieval decisions evolve during GRPO, we analyze the eight direct-or-retrieval choices within each logged GRPO group for MIntRec2.0, where all eight trajectories are sampled for the same training context. For a group with $k$ retrieval trajectories, the binary entropy $H_2(k/8)$ is zero when all trajectories select the same path and one bit when the group is evenly split. We divide the ordered training steps into early, middle, and late thirds, exclude groups containing format errors, and construct the reference by independently sampling eight path choices using IntPro's observed retrieval rate within each 100-step block.

Figure \ref{fig:grpo_decision_concentration} shows that IntPro's aggregate retrieval rate declines modestly across training, from 18.45\% in the early phase to 16.91\% and 16.65\% in the middle and late phases. Over the same phases, mean group entropy decreases from 0.434 to 0.345 bits and remains below the rate-matched reference range of 0.539--0.571 bits. Consistently, the share of all-direct groups increases from 41.1\% to 52.1\%, while groups containing one to three retrieval trajectories decrease from 44.6\% to 33.4\%; groups containing four to eight retrieval trajectories remain near 14.5\%. These results indicate greater within-group consistency in inference-path choices in later training phases, even as aggregate retrieval frequency changes only modestly.

\begin{figure*}[!t]
\centering
\includegraphics[width=\textwidth]{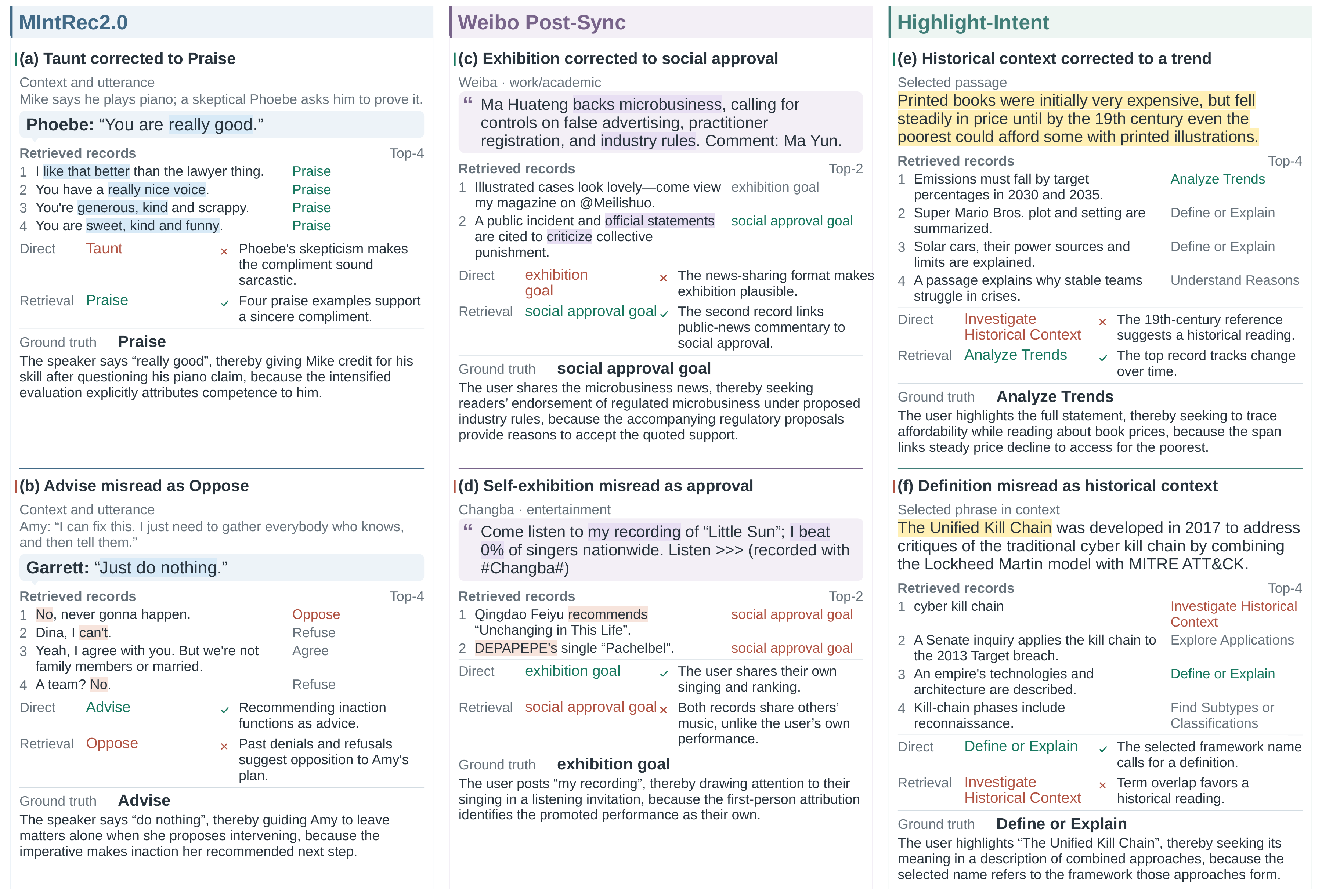}
\caption{Paired examples from IntPro with the Qwen3-4B backbone across MIntRec2.0, Weibo Post-Sync, and Highlight-Intent. In each dataset, retrieval corrects a direct error in the upper case and changes a correct direct prediction into an error in the lower case.}
\label{fig:case_study}
\end{figure*}

\subsubsection{Retrieval Budget Sensitivity}

We next test whether one retrieval budget is suitable across all three datasets. Table~\ref{tab:retrieval_budget} evaluates $K\in\{1,2,4\}$. The training-set selection gives $K=4$ for Highlight-Intent and MIntRec2.0 and $K=2$ for Weibo Post-Sync; the remaining rows show how alternative budgets transfer to the test set.

\begin{table}[t]
\centering
\caption{Retrieval-budget sensitivity of Selective Retrieval based on Qwen3-4B.}
\label{tab:retrieval_budget}
\begin{tabular}{llcc}
\toprule
\textbf{Budget} & \textbf{Dataset} & \textbf{Acc (\%)} & \textbf{M-F1 (\%)} \\
\midrule
\multirow{3}{*}{$K=1$}
& Highlight-Intent & 47.34 & 31.06 \\
& MIntRec2.0 & 66.43 & 58.79 \\
& Weibo Post-Sync & 79.22 & 63.88 \\
\addlinespace
\multirow{3}{*}{$K=2$}
& Highlight-Intent & 50.23 & 32.81 \\
& MIntRec2.0 & 68.27 & 59.04 \\
& Weibo Post-Sync & 79.63 & 64.83 \\
\addlinespace
\multirow{3}{*}{$K=4$}
& Highlight-Intent & 51.45 & 33.07 \\
& MIntRec2.0 & 69.70 & 59.24 \\
& Weibo Post-Sync & 78.60 & 61.95 \\
\bottomrule
\end{tabular}
\end{table}

Increasing $K$ from 1 to 4 steadily improves Accuracy and Macro-F1 on Highlight-Intent and MIntRec2.0. Weibo Post-Sync instead peaks at $K=2$ and declines at $K=4$. The selected budget therefore varies with the dataset rather than following a single rule across scenarios.

\subsection{Class-Level Analysis}

Dataset-level aggregate scores can hide how gains vary with intent frequency. Figure~\ref{histogram_class} therefore compares the two least frequent and two most frequent MIntRec2.0 intents for the four Qwen3-4B student variants and Qwen3-30B-A3B under direct inference. IntPro has higher Accuracy than Answer-only SFT on all four displayed categories, with the largest difference on the rare ``Emphasize'' category. The gain over Answer-only SFT thus appears in both rare and frequent intents within this four-class analysis. The relative advantage is not uniform across methods: Naive GRPO remains stronger on the frequent ``Inform'' category, while Qwen3-30B-A3B is competitive on ``Explain.''

\subsection{Case Study}

Finally, Figure~\ref{fig:case_study} presents, for each dataset, one case in which retrieval corrects an incorrect direct prediction and another in which it turns a correct direct prediction into an error. On MIntRec2.0, four prior \textit{Praise} records resolve a potentially sarcastic reading of ``You are really good,'' whereas denial-heavy records shift the advisory imperative ``Just do nothing'' toward \textit{Oppose}. On Weibo Post-Sync, a public-news record with an embedded stance supports \textit{Social Approval} for the microbusiness post, while two records about sharing others' music misdirect the model on a post displaying the user's own recording. On Highlight-Intent, the top record centered on change over time redirects the book-price passage to \textit{Analyze Trends}, whereas the near-duplicate term ``cyber kill chain'' shifts a definition request toward \textit{Investigate Historical Context}. These cases illustrate why the retrieval decision must be context dependent: retrieved records help when their functional relation transfers to the current action and can mislead when lexical or topical similarity dominates that relation.

\section{Conclusions}
\label{sec:conclusion}
We study context-aware intent inference when observed behavior may not explicitly reveal the user's goal and the current context alone may be insufficient. IntPro represents historical records with label-conditioned intent explanations and forms retrieval queries with candidate intent explanations generated from the current context, while its retrieval-decision reward trains the proxy to decide whether to invoke retrieval. Across three scenarios and three model backbones, IntPro obtains higher Accuracy, Macro-F1, and Weighted-F1 than Naive GRPO in every combination of dataset and model backbone. The intent explanation also achieves the highest same-intent retrieval results as both a library representation and a test-time query, and the Qwen3-4B analysis shows that Selective Retrieval exceeds fixed No Retrieval and Always Retrieval policies. These results support combining intent-aligned retrieval with a context-dependent retrieval decision for context-aware intent inference under the evaluated per-user library setting.

\section{Limitations}
\label{sec:limitations}
Our evaluation focuses on intent inference and retrieval by the proxy, so the reported results do not establish the quality of downstream LLM responses generated from the structured intent. IntPro is evaluated with fixed per-user intent libraries whose historical intent explanations are generated by the teacher model from annotated intent labels. The evidence is therefore bounded to the dataset-specific identity units and supervised historical libraries represented by MIntRec2.0, Weibo Post-Sync, and Highlight-Intent; performance with unlabeled or continually evolving user histories remains outside the evaluated setting. The remaining 5.97--6.79-point gap to the path oracle also shows that retrieval decisions are still imperfect under the same paired outputs.

% \section*{Acknowledgment}
% This research was sponsored by CAAI-Lenovo Blue Sky Research Fund and National Natural Science Foundation of China (NSFC) under the Grant No. 62372113. Peng Zhang is a faculty of College of Computer Science and Artificial Intelligence, Fudan University. Tun Lu is a faculty of College of Computer Science and Artificial Intelligence, School of International Communication and Global Leadership, Fudan University, and Shanghai Key Laboratory of Data Science.

% main.bbl is shipped; arXiv does not need to run BibTeX.
\bibliographystyle{IEEEtran}
\bibliography{intpro_refs}

\end{document}